\documentclass[10pt,twocolumn,letterpaper]{article}

\usepackage{cvpr}              

\usepackage{graphicx}
\usepackage{amsmath}
\usepackage{amssymb}
\usepackage{booktabs}

\usepackage{color}

\usepackage[pagebackref,breaklinks,colorlinks]{hyperref}

\usepackage[capitalize]{cleveref}
\crefname{section}{Sec.}{Secs.}
\Crefname{section}{Section}{Sections}
\Crefname{table}{Table}{Tables}
\crefname{table}{Tab.}{Tabs.}

\def\cvprPaperID{6647} 
\def\confName{CVPR}
\def\confYear{2023}

\begin{document}

\title{MeMaHand: Exploiting Mesh-Mano Interaction \\
for Single Image Two-Hand Reconstruction}


\author{
Congyi Wang$^{\ast}$~~~~
Feida Zhu$^{\ast}$~~~~
Shilei Wen$^{\dagger}$~~~~ \\
ByteDance\\
{\tt\small \{kongyi.1990, zhufeida, zhengmin.666\}@bytedance.com} \\
}

\maketitle

{
\renewcommand{\thefootnote}{\fnsymbol{footnote}}
\footnotetext[1]{~Equal contribution. $^{\dagger}$ Corresponding author.}
}

\begin{abstract}
Existing methods proposed for hand reconstruction tasks usually parameterize a generic 3D hand model or predict hand mesh positions directly. The parametric representations consisting of hand shapes and rotational poses are more stable, while the non-parametric methods can predict more accurate mesh positions. In this paper, we propose to reconstruct meshes and estimate MANO parameters of two hands from a single RGB image simultaneously to utilize the merits of two kinds of hand representations. To fulfill this target, we propose novel Mesh-Mano interaction blocks (MMIBs), which take mesh vertices positions and MANO parameters as two kinds of query tokens. MMIB consists of one graph residual block to aggregate local information and two transformer encoders to model long-range dependencies. The transformer encoders are equipped with different asymmetric attention masks to model the intra-hand and inter-hand attention, respectively. Moreover, we introduce the mesh alignment refinement module to further enhance the mesh-image alignment. Extensive experiments on the InterHand2.6M benchmark demonstrate promising results over the state-of-the-art hand reconstruction methods.  
\end{abstract}

\begin{figure}
\centering
\includegraphics[width=0.9\hsize]{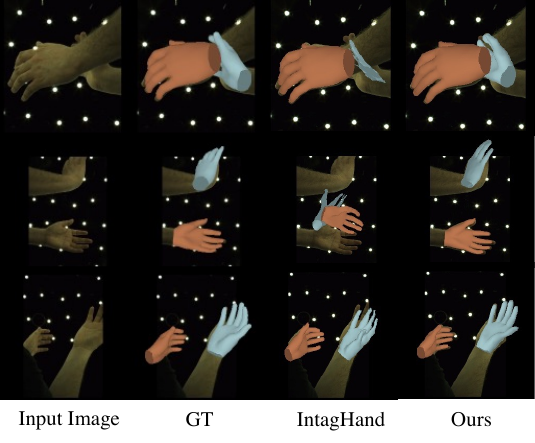}
\vspace{-3mm}
\caption{
Comparison with the state-of-the-art method IntagHand \cite{li2022interacting} for single-image two-hand reconstruction. The integration of parametric and non-parametric hand representations allows us to achieve better performance in hard cases such as severe occlusions and challenging viewpoints.
} 
\label{fig:teaser}
\end{figure}

\section{Introduction}

Vision-based 3D hand analysis plays an important role in many applications such as virtual reality (VR) and augmented reality (AR). Two-hand reconstruction from a single RGB image is more challenging due to complex mutual interactions and occlusions. Besides, the skin appearance similarity makes it difficult for the network to align image features to the corresponding hand.

Previous hand reconstruction works can be divided into two categories, parametric methods \cite{boukhayma20193d,zhang2019end,zhou2020monocular,chen2021model} and non-parametric methods \cite{cai2018weakly,ge20193d,moon2020i2l,chen2021i2uv,chen2021camera,lin2021end,lin2021mesh}. Parametric methods typically learn to regress \textit{pose} and \textit{shape} parameters of MANO model \cite{romero2022embodied}, where \textit{pose} represents joint rotations in axis-angle representation and \textit{shape} represents the coefficients of shape PCA bases. The MANO prior can yield plausible hand shapes from a single monocular image. However, they can not produce fine-grained hand meshes due to their limited capacity. 

With the rapid progress of graph convolutional network (GCN) and transformer techniques \cite{ge20193d,lin2021end,lin2021mesh, li2022interacting}, it is observed that direct mesh reconstruction can achieve state-of-the-art performance towards the Euclidean distances between the ground truth vertices and the predicted vertices. Nonetheless, the non-parametric methods are less robust in handling challenging viewpoints or severe occlusions. 

In this paper, we introduce a novel single-image two-hand reconstruction method designed to predict mesh vertices positions and estimate MANO parameters simultaneously to utilize the merits of two kinds of hand representations. The proposed \textbf{Me}sh-\textbf{Ma}no interaction \textbf{Hand} reconstruction architecture (MeMaHand) consists of three modules: 1) the image encoder-decoder module, 2) the mesh-mano interaction module, 3) and the mesh alignment refinement module. To extract contextually meaningful image features, we pre-train a classical image encoder-decoder network on auxiliary tasks including hand segmentation, hand 2D joints and dense mapping encodings. The low-resolution features encode more global knowledge, while the high-resolution features contain more local details. Secondly, the mesh-mano interaction module stacks three mesh-mano interaction blocks (MMIBs) to transform the mesh vertices and MANO parameters queries initialized by the global image feature vector. We observe that the hand prior embedded in the MANO parameters is valuable for predicting stable hand meshes in challenging situations such as severe occlusions. MMIB consists of one graph residual block to aggregate local information and two transformer encoders to model long-range dependencies. The transformer encoders are equipped with different asymmetric attention masks to model the intra-hand and inter-hand attention, respectively. Each MMIB is followed by an upsampling operation to upsample the mesh vertices tokens in a coarse-to-fine manner. Finally, the mesh alignment refinement module utilizes one MMIB to predict offsets for mesh vertices and MANO parameters to enhance mesh-image alignment. To improve the reliability of image evidence, we project mesh vertices predicted by the Mesh-Mano interaction module onto the 2D image plane. The explicit mesh-aligned image features are concatenated to the transformer input tokens.

\begin{figure*}[!t]
\centering
\includegraphics[width=0.98\textwidth]{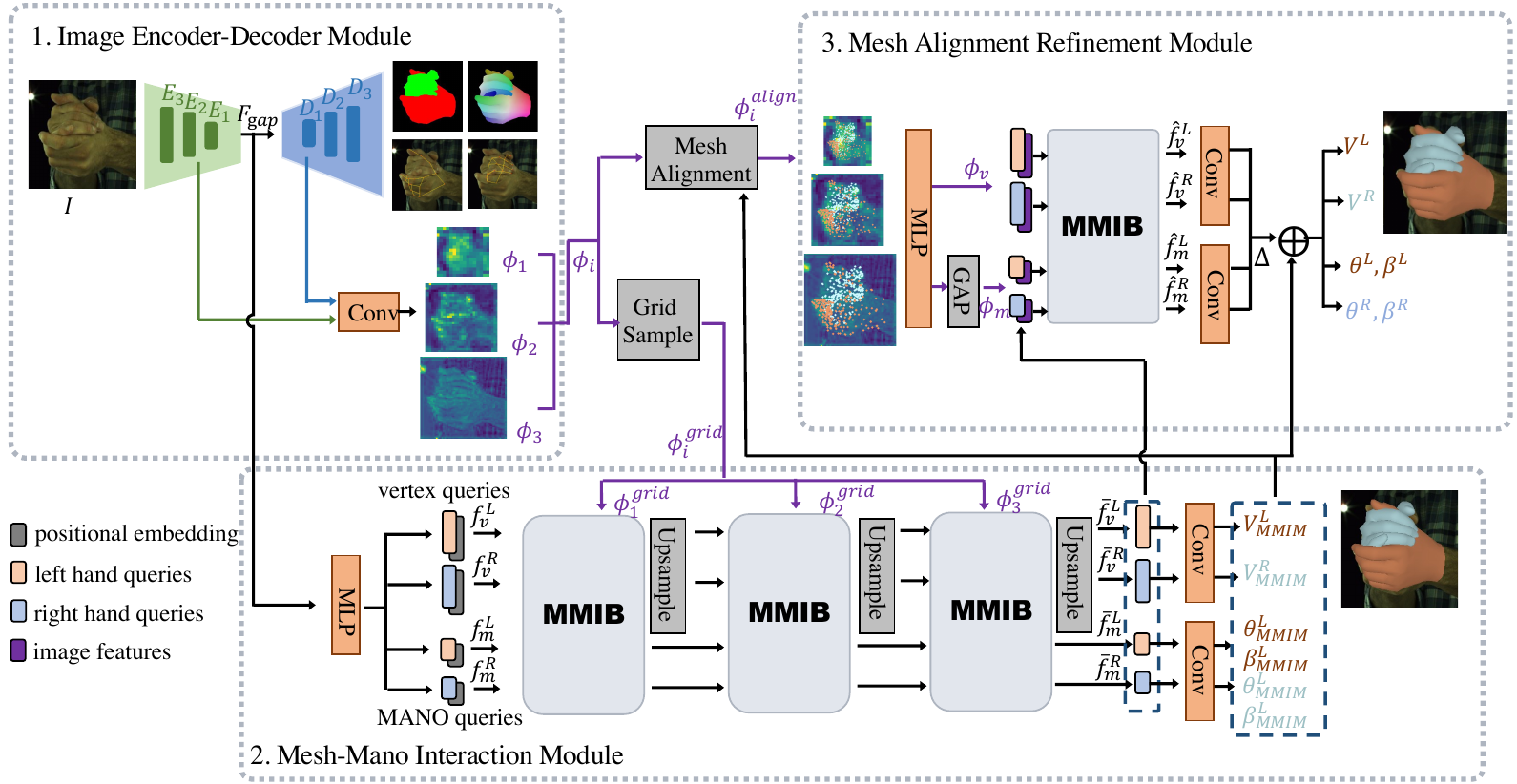}
\caption{The overall architecture of our proposed MeMaHand. We pretrain an image encoder-decoder module on auxiliary tasks to extract multi-scale contextually meaningful image features. The Mesh-Mano interaction module stacks three Mesh-Mano interaction blocks (MMIBs) to predict the mesh vertices positions and MANO parameters simultaneously. The mesh alignment refinement module rectifies the mesh-image misalignment by explicitly utilizing the mesh-align image features.}
\label{fig:architecture}
\end{figure*}

The whole network, including the pre-trained image feature encoder-decoder, is jointly optimized such that the image features better adapt to our hand mesh reconstruction task. Benefiting from the mesh-mano interaction mechanism and mesh alignment refinement stage, extensive experiments demonstrate that our method outperforms existing both parametric and non-parametric methods on InterHand2.6M \cite{moon2020interhand2} dataset. In summary, the contributions of our approach are as follows: 

\vspace{-1mm}
\begin{itemize}
    \setlength{\itemsep}{0pt}
    \setlength{\parsep}{0pt}
    \setlength{\parskip}{0.2pt}
    \item We propose MeMaHand to integrate the merits of parametric and non-parametric hand representation. The mesh vertices and MANO parameters are mutually reinforced to achieve better performance in mesh recovery and parameter regression.
    \item A mesh-image alignment feedback loop is utilized to improve the reliability of image evidence. Therefore, more accurate predictions are obtained by rectifying the mesh-image misalignment.
    \item Our method achieves superior performance on the InterHand2.6M dataset, compared with both non-parametric and parametric methods. 
\end{itemize}

\section{Related Works}

\noindent \textbf{Parametric Hand Reconstruction.}
Parametric approaches \cite{baek2019pushing,baek2020weakly,hasson2020leveraging,hasson2019learning,zhang2019end,romero2022embodied,boukhayma20193d,zhou2020monocular,chen2021model} use a parametric hand model such as MANO \cite{romero2022embodied} and focus on regressing the \textit{pose} and \textit{shape} parameters from a single image. The rich embedded prior information (\emph{e.g.} the geometric dynamic constraints of joint rotations) can assist deep model learning when 3D annotations are insufficient. Weak 2D joints supervision \cite{boukhayma20193d} and motion caption data \cite{zhou2020monocular} are utilized to train convolutional neural networks (CNN) to predict MANO parameters. The reliance on 3D manual annotations is further totally alleviated by S$^{2}$HAND \cite{chen2021model}. However, the reconstructed mesh can not fully express the local details of variable 3D hand shapes due to the limited capacity of the MANO model. 


\vspace{1mm}
\noindent \textbf{Non-parametric Hand Reconstruction.} Non-parametric methods aim to reconstruct the hand mesh directly \cite{ge20193d,moon2020i2l,chen2021i2uv,chen2021camera,Tang_2021_ICCV,lin2021end, li2022interacting,lim2020mobilehand,lin2021mesh,Moon_2020_ECCV_I2L-MeshNet,choi2020pose2mesh,wang2018pixel2mesh}. To explicitly encode mesh topology, graph convolutional network (GCN) is adopted for aggregating adjacent vertices features. Hierarchical architectures \cite{ge20193d,li2022interacting} are designed for mesh generation from coarse to fine. To model long-range dependencies, multi-layer transformer encoders are introduced such that the global interactions can be modeled without being limited by any mesh topology \cite{lin2021end,lin2021mesh}. Most recently, IntagHand \cite{li2022interacting} achieves the state-of-the-art performance on InterHand2.6M dataset \cite{moon2020interhand2}. However, we observe that IntagHand is less robust in handling challenging viewpoints or severe occlusions. The reconstructed mesh may be corrupted into unnatural shapes.

\vspace{1mm}
\noindent \textbf{Interacting Two-Hand Reconstruction.} Although single-hand methods can extend to two-hand reconstruction, the correlation between the left and right hands is not considered. Besides, the performance deteriorates for the close-interacting hands. Depth cameras \cite{kyriazis2014scalable} are sensitive to tracking accuracy and multi-view images \cite{tzionas2016capturing,smith2020constraining} are expensive to acquire. 
Based on a large-scale interacting hand dataset named InterHand2.6M \cite{moon2020interhand2},  deep learning based single image methods either estimate hand joint positions \cite{moon2020interhand2,kim2021end}, or employ a 2.5D heatmap to predict the MANO parameters \cite{zhang2021interacting}, or directly reconstruct meshes \cite{li2022interacting}. In summary, existing methods only employ a single specific hand representation. In contrast, our method not only combines the merits of MANO representations and mesh representations but also proposes a dedicated transformer encoder with asymmetric attention masks to make them collaborate seamlessly. 
Besides, the joint rotation and precise mesh vertices positions can be used for different applications, such as driving CG characters or virtual try-on.





\section{Methodology}
The overall architecture of MeMaHand is depicted in \cref{fig:architecture}. Given a single image $I$, the proposed method can predict the mesh vertices positions $V \in  \mathbb{R}^{778 \times 3}$, MANO pose parameter $\theta \in \mathbb{R}^{48}$ and shape parameter $\beta \in \mathbb{R}^{10}$ of two hands simultaneously in one forward pass. In this section, we first describe the overall architecture of MeMaHand in \cref{sec:system overview}. Then, we illustrate the mesh-mano interaction module and mesh-alignment refinement module in \cref{sec:Mesh-Mano Interaction Module} and \cref{sec:Mesh-alignment Refinement Module}, respectively. Finally, we introduce the model objectives used to train our network in \cref{sec:model objectives}.

\subsection{System Overview}
\label{sec:system overview}

To extract contextually meaningful image features, we pre-train a classical image encoder-decoder module on auxiliary tasks including hand segmentation, hand 2D joints, and dense mapping encodings. The 2D image conditions are generated from the reconstructed mesh of ground-truth MANO parameters. ResNet50~\cite{he2016deep} is leveraged as the backbone. Global average pooling feature $F_{gap}$ and multi-scaled image feature $\{\phi_{i} \in \mathbb{R}^{C_{i} \times H_{i} \times W_{i}},i=1,2,3 \}$ are extracted from the image encoder (E) and decoder (D).

Afterward, we propose Mesh-Mano Interaction Module to predict mesh vertices positions $V_{MMIM}$ and MANO parameters $\theta_{MMIM}, \beta_{MMIM}$ of both left (L) and right (R) hands. We use the global feature $F_{gap}$ to initialize the vertex and MANO queries. The grid-sampled image features $\{\phi_{i}^{grid},i=1,2,3\}$ are tokenized as well following the practice of Mesh Graphformer \cite{lin2021mesh}. 

The mesh alignment refinement module further improves the predictions generated by the mesh-mano interaction module. The explicit mesh-aligned image features extracted from the multi-scale feature maps $\phi_{i}$ are more informative than grid-sampled image features, which helps rectify the predictions. Details will be elaborated in \cref{sec:Mesh-alignment Refinement Module}.

\subsection{Mesh-Mano Interaction Module}
\label{sec:Mesh-Mano Interaction Module}

The mesh-mano interaction module reconstructs hand mesh in a coarse-to-fine manner with three Mesh-Mano Interaction Blocks (MMIBs). We leverage the graph coarsening method \cite{li2022interacting} to build three-level sub meshes with vertex number $N_{1}=63,N_{2}=126,N_{3}=252$. Each MMIB is followed by an upsample operation that reverses the topological relationship between adjacent sub-meshes. The full mesh vertices positions ($N=778$) are obtained with a simple $1\times1$ $Conv$ from the final output vertex token.

\begin{figure}[!t]
\centering
\includegraphics[width=0.4\textwidth]{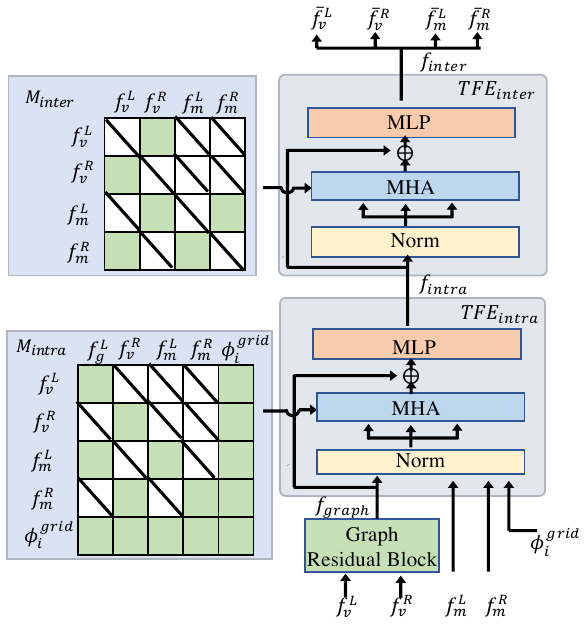}
\vspace{-1mm}
\caption{The detailed structure of Mesh-Mano Interaction Block (MMIB). MMIB consists of one graph residual block to aggregate local information and two transformer encoders to model long-range dependencies.}
\label{fig:MMIB}
\end{figure}

\cref{fig:MMIB} shows the detailed structure of MMIB. At level $i$, the input tokens of MMIB include two kinds of queries: vertex queries $f_v^h \in \mathbb{R}^{N_{i} \times D_{i}}$ and MANO parameter queries $f_m^{h} \in \mathbb{R}^{2 \times D_{i}}$, where $h$ indicates left ($L$) or right ($R$) hand, and $D_i$ represents the feature dimensions. Therefore, the total sequence length is $(2 \times N_i + 4)$.

\vspace{1mm}
\noindent \textbf{Graph Residual Block.} The design of the graph residual block is similar to \cite{ge20193d,li2022interacting}. The Chebyshev spectral graph CNN \cite{defferrard2016convolutional} is adopted to transform the vertex token to intermediate graph features,
\begin{equation}
\label{eq:gcn}
f_{graph} = \mathbf{GraphConv}(f_v^L, f_v^R),
\end{equation}
where the operation $\mathbf{GraphConv}$ denotes the graph convolution. More details can be found in \cite{defferrard2016convolutional}. For simplicity, we use $f_{graph}$ to represent the concatenation of updated left-hand and right-hand vertex tokens. 

\vspace{1mm}
\noindent \textbf{Intra-hand Transformer Encoder.} While graph CNN is useful for extracting local information, it is less efficient at capturing long-range dependencies. We use transformer encoders to model dependencies not only between long-range vertices but also between mesh vertices and MANO parameters. The spatial image features at resolution $i$ are also tokenized as $\phi_{i}^{grid}$ by grid sampling following the practice of Mesh Graphformer \cite{lin2021mesh}. Therefore, the input tokens of intra-hand transformer encoder ($TFE_{intra}$) consist of graph features ($f_{graph}$), MANO queries ($f_m^L,f_m^R$) and grid image feature ($\phi_{i}^{grid}$). 
Based on the observation that the mesh vertices regression can lead to more precise mesh reconstruction and the MANO parameters are more stable, we propose an asymmetric attention mask excluding the mano-to-mesh attention such that the MANO parameter tokens will not directly affect the mesh vertex tokens. The mesh-to-mano attention remains. Therefore, the mesh vertex tokens are implicitly regularized by the MANO parameters loss. $TFE_{intra}$ 
focuses on modeling the intra-hand dependencies. The inter-hand attention is also excluded. The resulting asymmetric attention mask $M_{intra}$ is shown in \cref{fig:MMIB}. Finally, the intra-hand transformer encoder is formulated as:
\begin{equation}
\label{eq:mesh_mano_intra}
f_{intra} = TFE_{intra}([f_{graph},f_m^L, f_m^R, \phi_i^{grid}], M_{intra}),
\end{equation}
where $[...]$ denotes the concatenation operation. For simplicity, $f_{intra}$ represents the concatenation of updated mesh vertex and MANO parameter tokens.

\vspace{1mm}
\noindent \textbf{Inter-hand Transformer Encoder.}  In contrast to $TFE_{intra}$, the inter-hand transformer encoder ($TFE_{inter}$) focuses on modeling the inter-hand correlations.  \cref{fig:MMIB} presents the resulting asymmetric attention mask $M_{inter}$. The image feature tokens are not used in $TFE_{inter}$. Finally, the inter-hand transformer encoder is formulated as:
\begin{equation}
\label{eq:mesh_mano_inter}
f_{inter} = TFE_{inter}(f_{intra}, M_{inter}),
\end{equation}
where $f_{inter}$ can be further split into updated vertex tokens $\bar f_v^L, \bar f_v^R$ and MANO parameter tokens  $\bar f_m^L, \bar f_m^R$.

Our proposed MMIB predicts the mesh vertices positions and MANO parameters simultaneously in a unified architecture. Intra-hand and inter-hand mesh-mano dependencies are modeled by two cascaded transformer encoders with different attention masks. After three MMIBs, a simple $1 \times 1$ $Conv$ is applied to obtain the mesh vertices positions and MANO parameters from output tokens,
\begin{align}
\label{eq:MMIB_1_1_conv_vertex} V_{MMIM}^h = & Conv_{v}(\bar f_{v}^h), \\
\label{eq:MMIB_1_1_conv_MANO} \theta_{MMIM}^h, \beta_{MMIM}^h = & Conv_{m}(\bar f_{m}^h),
\end{align}
where $h \in \{L,R\}$ indicates the left or right hand. To tackle the noticeable misalignment between the estimated meshes and image evidence, we propose a mesh alignment refinement module to rectify the results. 


\subsection{Mesh Alignment Refinement Module}
\label{sec:Mesh-alignment Refinement Module}

To further improve mesh-image alignment, we propose a novel mesh-alignment refinement module inspired by PyMAF \cite{zhang2021pymaf}. Specifically, we project the mesh vertices $V_{MMIM}$ predicted by the mesh-mano interaction module onto the multi-scale image features explicitly. Three-scale mesh-aligned image features 
are concatenated together for each corresponding vertex token. A simple multi-layer perceptron (MLP) is adopted to reduce the dimensions, resulting in a fused image feature $\phi_{v}$. 

We utilize one MMIB to refine the mesh vertex and MANO parameter tokens $\bar f_v^h, \bar f_m^h$. To effectively utilize the mesh-aligned image evidence, we made some modifications to the tokens. Specifically, we concatenate the image feature $\phi_{v}$ to vertex token $\bar f_v^h$ along the channel dimension. For MANO parameter tokens, we perform a global average pooling operation to obtain global feature vector $\phi_{m}$ before concatenation. The rectified tokens are formulated as,
\begin{equation}
\hat f_v^h,  \hat f_m^h  = MMIB(\bar f_v^h, \bar f_m^h, \phi_v, \phi_m).
\end{equation}

One simple $1 \times 1$ $Conv$ layer is applied to obtain the mesh vertices positions offsets and MANO parameter offsets for rectification,
\begin{align}
 V^h = & V_{MMIM}^h + Conv_{v}(\hat f_{v}^h), \\
\theta^h  = & \theta_{MMIM}^h + Conv_{m}(\hat f_{m}^h), \\
\beta^h  = & \beta_{MMIM}^h + Conv_{m}(\hat f_{m}^h),
\end{align}
where $V^h, \theta^h$ and $\beta^h$ are the final results. Note that we can stack several MMIBs to predict the offsets iteratively. In our experiments, one single MMIB is enough to achieve satisfactory results. 
PyMAF \cite{zhang2021pymaf} fuses the mesh-align features into one global feature, which ignores the inherent spatial relationship of mesh vertices. In contrast, our MMIB adopts GCN to model the spatial relation between adjacent mesh vertices. The transformer encoder is responsible for modeling the long-range dependencies.

\begin{figure*}[!t]
\centering
\includegraphics[width=0.99\textwidth]{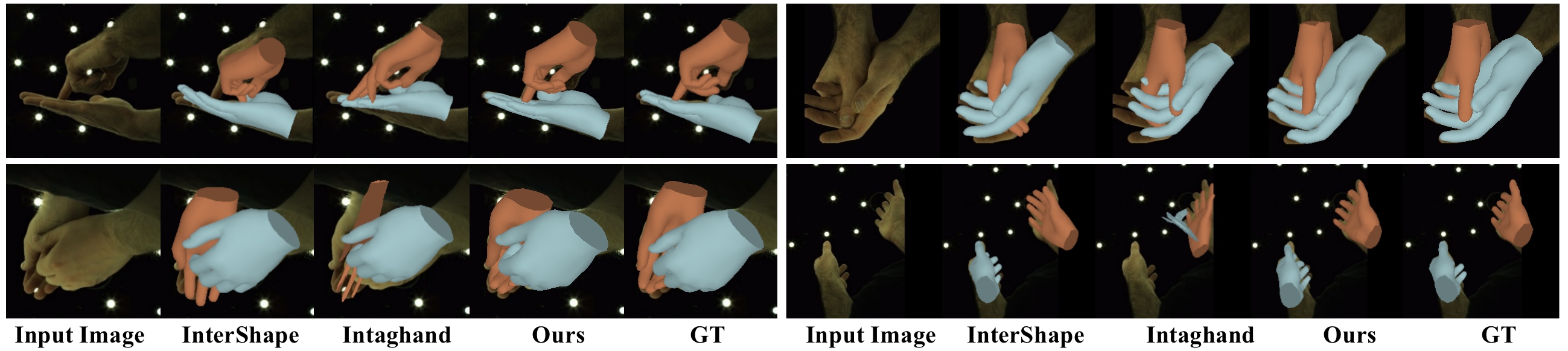}
\caption{Qualitative comparison with SOTA parametric method InterShape \cite{zhang2021interacting} and SOTA non-parametric method IntagHand \cite{li2022interacting} on InterHand2.6M dataset. Our method performs better on close-interacting two-hand reconstruction (first row). Besides, our method is robust in hard cases such as severe occlusions and challenging viewpoints (second row).}
\label{fig:compare}\textbf{}
\end{figure*}

\begin{figure*}[!t]
\centering
\includegraphics[width=0.99\textwidth]{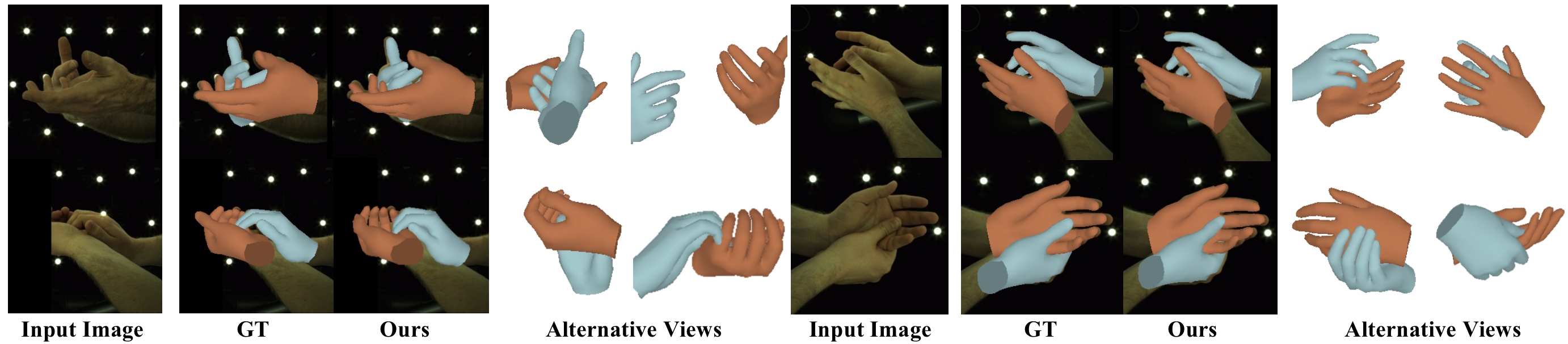}
\caption{Our method produces accurate hand reconstructions for various two-hand interactions.}
\label{fig:fig_our}
\end{figure*}

\subsection{Model Objectives}
\label{sec:model objectives}

Our method predicts mesh vertices positions and MANO parameters simultaneously. The learning objectives can be divided into three categories: mesh vertex loss, MANO parameter loss and mesh-mano consistency loss.


\vspace{1mm}
\noindent \textbf{Mesh Vertex Loss:} The widely-used L1 loss is adopted to supervise the vertex positions and the 2D projections:
\begin{equation}
\mathcal L_{V} =  \sum_{h=L,R}\|V^h - V_{GT}^h\|_{1} + \|\Pi(V_h) - \Pi(V_{GT}^h)\|_{1},
\end{equation}
where $V_{GT}^h$ represents the ground-truth vertex positions and $\Pi$ denotes the projection function. Given vertex positions, the joint positions can be regressed by multiplying the predefined regression matrix $\mathcal{J}$. Joint losses are formulated as:
\begin{equation}
\begin{split}
    \mathcal L_{J} = &\sum_{h=L,R}\|\mathcal{J} V^h - \mathcal{J} V_{GT}^h\|_{1} \\ 
                       + & \sum_{h=L,R}\|\Pi\mathcal{J} (V_h) - \Pi(\mathcal{J} V_{GT}^h)\|_{1}.
\end{split}
\end{equation}
 
The face normal loss is introduced to regularize the surface normal consistency:
\begin{equation}
\mathcal L_{N} = \sum_{h=L,R} \sum_{f=1}^{F}\sum_{i=1}^{3} \|e_{f,i}^h \cdot n_{f,GT}^{h}\|_{1},
\end{equation}
where $e_{f,i}^h$ represents the $i$th edge of face $f$ at hand $h$ and $n_{f,GT}^{h}$ is the normal vector of this face from the ground truth mesh. The edge length consistency loss is to enforce the edge length consistency:
\begin{equation}
\mathcal L_{E} = \sum_{h=L,R} \sum_{i=1}^{E} \|e_{i}^h - e_{i,GT}^{h}\|_{1},
\end{equation}
where $e_{i}^h$ represents the $i$th edge of hand $h$ and $E$ denotes the total edge number, respectively.

\vspace{1mm}
\noindent \textbf{MANO Parameter Loss:} Given the ground truth MANO parameters, L1 loss is used to regress parameters,
\begin{equation}
\mathcal L_{P} = \sum_{h=L,R} \|\theta^h - \theta_{GT}^{h}\|_{1}+\|\beta^h - \beta_{GT}^{h}\|_{1}.
\end{equation}
In addition, we reconstruct hand mesh based on the MANO model. The reconstructed mesh should be close to the ground-truth mesh,  
\begin{equation}
\mathcal L_{V,mano} =  \sum_{h=L,R}\|\mathcal{MANO}(\theta^h,\beta^h) - V_{GT}^h\|_{1}.
\end{equation}

\vspace{1mm}
\noindent \textbf{Mesh-Mano Consistency Loss.} The predicted mesh and the reconstructed mesh from MANO parameters should be consistent with each other:
\begin{equation}
\begin{split}
\mathcal{L}_{consist} = & \sum_{h=L,R}\|\mathcal{MANO}(\theta^h,\beta^h) - V^h\|_{1} \\
+& \|\mathcal{J}\mathcal{MANO}(\theta^h,\beta^h) - \mathcal{J}V^h\|_{1}.  
\end{split}
\end{equation}

In summary, the overall training loss $\mathcal{L}$ is: 
\begin{equation}
\begin{split}
    \label{eq:loss_total} \mathcal{L} = & \lambda_{V} \mathcal{L}_{V} + \lambda_{J} \mathcal{L}_{J} +  \lambda_{N}\mathcal{L}_{N} +  \lambda_{E}\mathcal{L}_{E} \\
    +&  \lambda_{P}\mathcal{L}_{P} + \lambda_{V,mano}\mathcal{L}_{V,mano} + \lambda_{consist}\mathcal{L}_{consist}.
\end{split}
\end{equation}
where $\lambda_{V}=40$, $\lambda_{J}=40$, $\lambda_{N}=5$, $\lambda_{E}=40$, $\lambda_{P}=10$, $\lambda_{V,mano}=10$ and $\lambda_{consist}=40$. The whole network including the pre-trained image feature encoder-decoder is jointly optimized such that the image features better adapt to our hand mesh reconstruction task.

\begin{table}[!t]
\centering
\resizebox{0.4\textwidth}{!}
{
\renewcommand{\arraystretch}{1.1}
\begin{tabular}{c|cc}
\specialrule{.1em}{0em}{0em} 
Method       & Params(M) & Inference Time (ms) \\ \hline
InterShape \cite{zhang2021interacting}   & 139.6 & 47.77     \\ 
Intaghand \cite{li2022interacting}   & 37.28            & 51.86  \\ \hline
 Ours        & 38.31  & 60.52 \\  
\specialrule{.1em}{0em}{0em}
\end{tabular}
}
\caption{Parameters and inference speed on Tesla V100}
\label{table:time_compare}             
\end{table}

\section{Experiment}

\subsection{Datasets and Implementation}

\vspace{1mm}
\noindent \textbf{Training Dataset.} Interhand2.6M \cite{moon2020interhand2} is a large-scale hand dataset with ground truth mesh annotations including both single-hand and interacting two-hand images. We pick out the interacting hand (IH) data for training and testing. For a fair comparison, we follow the preprocessing steps of \cite{li2022interacting} which produces 366K training and 261K testing samples.

\vspace{1mm}
\noindent \textbf{Implementation Details.} The image encoder uses ResNet50 \cite{he2016deep} as the backbone. The decoder contains four simple deconvolutional layers. We first pretrain our image encoder-decoder module on auxiliary tasks. Then, the whole network is jointly optimized with Adam optimizer at a learning rate $1\times10^{-4}$. Data augmentation includes random rotation, random translation, and random scaling. The batch size is set to 32. It takes around 48 hours to train MeMaHand with 4 Tesla V100 GPUs. 

\vspace{1mm}
\noindent \textbf{Metrics.} Mean Per Joint Position Error (MPJPE) and Mean Per Vertex Position Error (MPVPE) in millimeters are adopted to evaluate the mesh reconstruction accuracy. Additionally, we report the percentage of correct keypoints (PCK) curve and Area Under the Curve (AUC) across the thresholds between 0 and 50 millimeters. To evaluate the mesh-image alignment accuracy, the reconstructed mesh vertices are projected onto the 2D image plane. The mesh-image alignment accuracy is measured by PROJ2D, which calculates the distance in image pixels between the projected ground truth vertices and the predicted vertices.


\newcommand{\FigIntuitiveWidth}{0.5}
\begin{figure*}[!t]{}
\captionsetup[subfigure]{labelformat=empty}
\centering
\subcaptionbox{(a) Comparison with non-parametric method}{\includegraphics[width=0.44\textwidth]{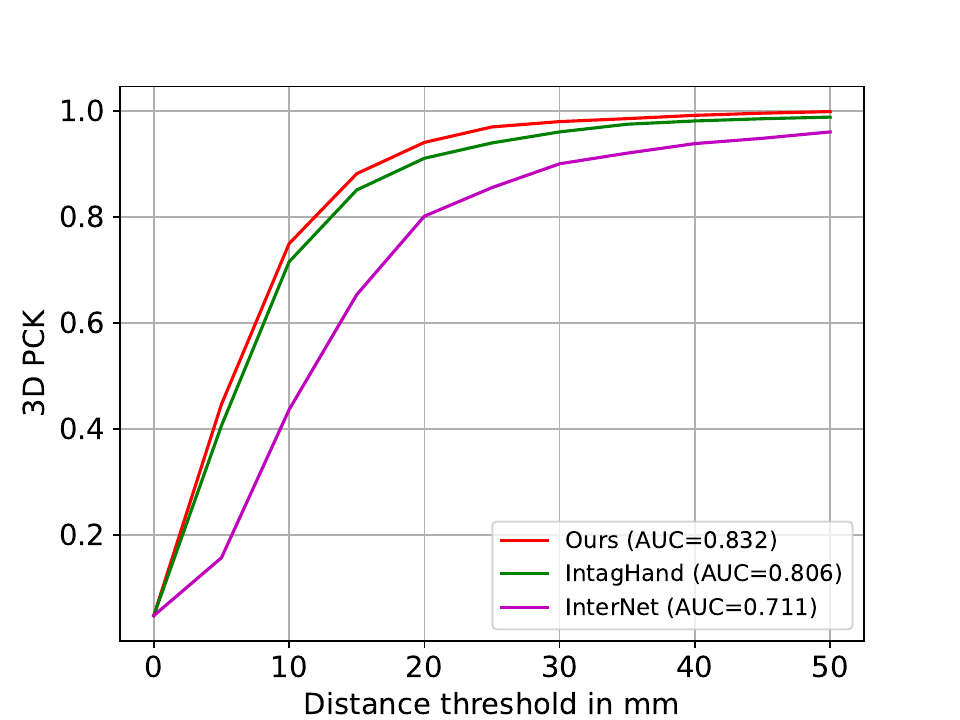}}
\hspace{0.03\textwidth}
\subcaptionbox{(b) Comparison with parametric method}{\includegraphics[width=0.44\textwidth]{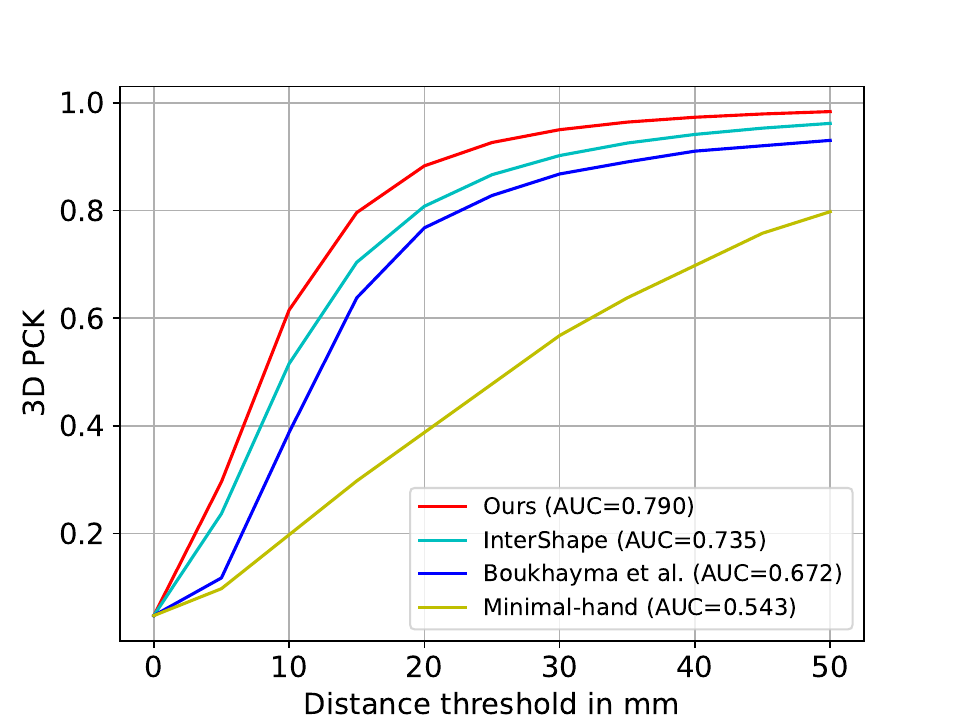}} 
\caption{Comparison with the state-of-the-art methods on InterHand2.6M dataset. \textbf{Left}: 3D PCK of non-parametric methods. \textbf{Right}: 3D PCK of parametric methods. The AUC values are shown in parentheses.}
\label{fig:auc_compare}
\end{figure*}

\begin{table}[!t]
\centering
\resizebox{0.46\textwidth}{!}
{
\renewcommand{\arraystretch}{1.2}
\begin{tabular}{c|cccc}
\specialrule{.1em}{0em}{0em}
Method       & MPJPE$\downarrow$ & MPVPE$\downarrow$ &   AUC $\uparrow$ & PROJ2D$\downarrow$ \\ \hline
zimmermann \etal \cite{zimmermann2017learning}   &36.36 &- &- &-\\ 
Spurr \etal \cite{spurr2018cross}               &15.40 &- &- &- \\
InterNet\cite{moon2020interhand2}            & 16.00 & -    & 0.711  & - \\ 
Digit\cite{fan2021learning} & 14.27 & -    & -  & - \\
Intaghand\cite{li2022interacting}                & 8.79  & 9.03 & 0.806 & 6.47 \\ \hline
 Ours                                & \textbf{8.65} & \textbf{8.89} &  \textbf{0.832}  & \textbf{6.22}  \\  
\specialrule{.1em}{0em}{0em}                       
\end{tabular}
}
\caption{Quantitative comparison with state-of-art non-parametric methods on InterHand2.6M dataset. 
}
\label{table:mesh_compare}             
\end{table}

\subsection{Comparison with State-of-the-art Methods}
We compare our model with state-of-the-art non-parametric methods including InterNet \cite{moon2020interhand2} and Intaghand\cite{li2022interacting} and parametric methods including \cite{boukhayma20193d}, MinimalHand \cite{zhou2020monocular} and InterShape\cite{zhang2021interacting}. The officially released weights are used to obtain the results. The model parameters and inference time of our method and other SOTA methods are reported in \cref{table:time_compare}. The inference of our model can be completed in $60.52ms$, comparable to other SOTAs but with better mesh reconstruction results. Besides, the mesh vertex positions as well as the MANO parameters are predicted simultaneously in one single forward pass.

\vspace{1mm}
\noindent \textbf{Comparison with Non-parametric Methods.} Non-parametric methods directly generate mesh vertice positions, which can express the local details of variable 3D hand shapes. The quantitative results of non-parametric methods are shown in \cref{table:mesh_compare}. It can be seen that our method achieves the best performance on all evaluation metrics. \cref{fig:auc_compare}(a) presents the PCK curve, which further demonstrates the superiority of our method at all threshold levels.

\cref{fig:compare} presents the qualitative comparison with SOTA non-parametric method IntagHand \cite{li2022interacting}. Our method performs better on close interacting two-hand reconstruction (first row). Besides, we observe that IntagHand produces collapsed meshes in hard cases such as severe occlusions and challenging viewpoints (second row). In contrast, our method is more robust to such situations. We attribute this success to the integration of MANO representations, which predicts stable hand meshes. More results with alternative views are shown in Figure \ref{fig:fig_our}.

\begin{table}[!t]
\centering
\resizebox{0.47\textwidth}{!}
{
\renewcommand{\arraystretch}{1.1}
\begin{tabular}{c|cccc}
\specialrule{.1em}{0em}{0em}
Method       & MPJPE$\downarrow$ & MPVPE $\downarrow$ & AUC$\uparrow$ & PROJ2D$\downarrow$ \\ \hline
Boukhayma \etal \cite{boukhayma20193d}   & 16.93             & 17.98    & 0.672          & - \\ 
MinimalHand\cite{zhou2020monocular}   & 23.48           & 23.89 & 0.543 & -\\ 
InterShape \cite{zhang2021interacting}   & 13.48           & 13.95 &0.735 &9.95     \\ \hline
 Ours        & \textbf{10.85}   & \textbf{10.92}  & \textbf{0.790}  &\textbf{6.90}\\  
\specialrule{.1em}{0em}{0em}                      
\end{tabular}
}
\caption{Quantitative comparison with state-of-art parametric methods on InterHand2.6M dataset. 
}
\label{table:mano_compare}             
\end{table}

\vspace{1mm}
\noindent \textbf{Comparison with Parametric Methods.} For a fair comparison, the reconstructed hand meshes from our predicted MANO parameters are utilized for evaluation. The quantitative comparisons are listed in \cref{table:mano_compare}. The PCK curve is shown in \cref{fig:auc_compare}(a). The MANO parameter estimation of our method outperforms other SOTA parametric methods. Thanks to the mesh-mano interaction block, the MANO token is conditioned on the mesh vertex tokens where spatial relations are retained. Such dependencies are more informative than one-dimensional global feature utilized in competing methods \cite{boukhayma20193d,zhou2020monocular,zhang2021interacting}.

From \cref{table:mesh_compare} and \cref{table:mano_compare}, we can find the parametric method InterShape \cite{zhang2021interacting} has higher statistical errors on joint and vertex positions compared with the non-parametric method IntagHand \cite{li2022interacting} due to the limited capacity of the parametric model. However, from the second row of \cref{fig:compare}, we can see InterShape produces reasonable hand shapes in these challenging cases. By combining the merits of parametric and non-parametric hand representations, our method achieves the best performance both quantitatively and qualitatively.


\begin{figure}[!t]
 \centering
\includegraphics[width=0.48\textwidth]{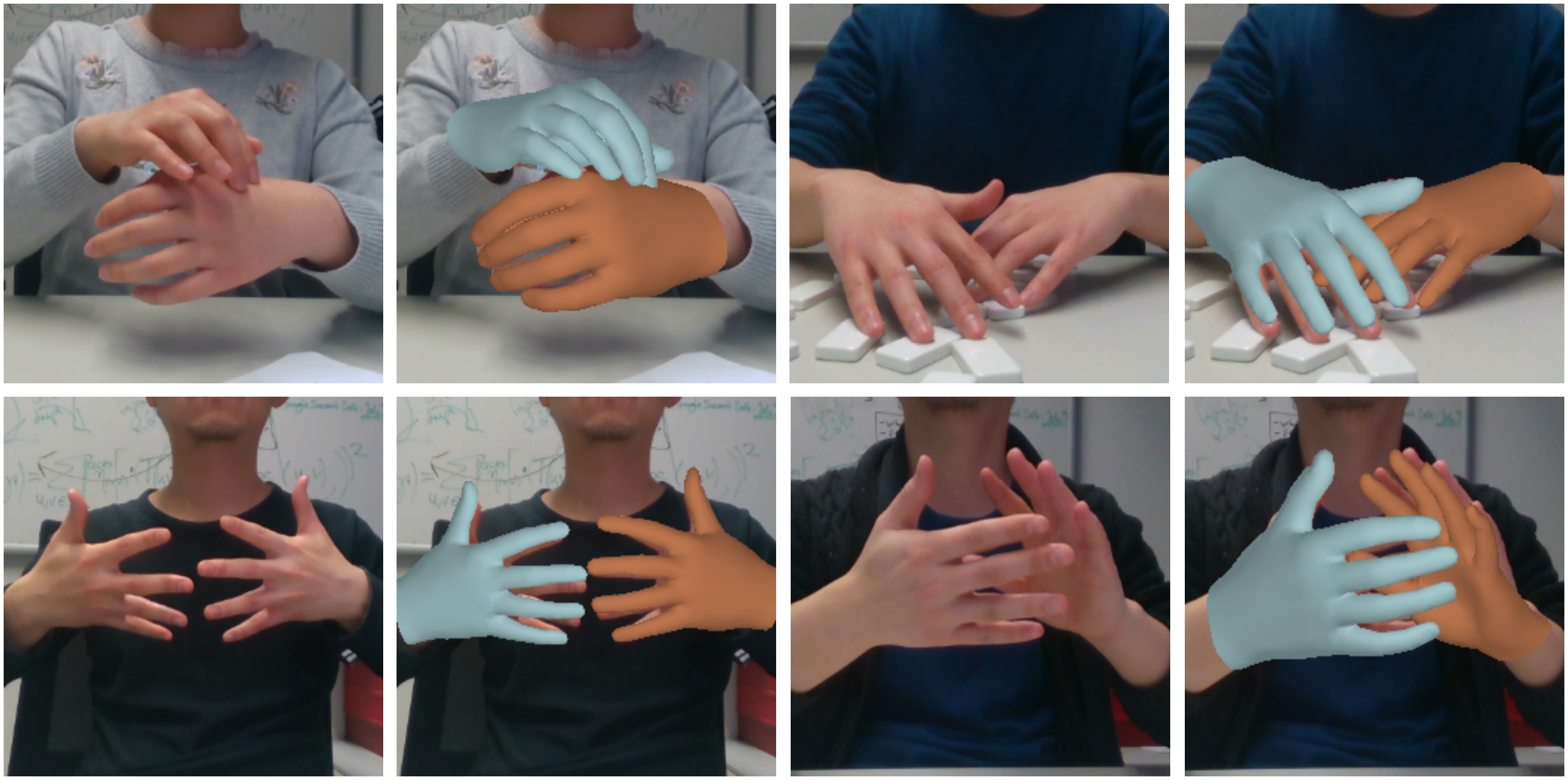}
 \caption{Two-hand reconstruction of images in the wild, which are taken from the RGB2Hands \cite{wang2020rgb2hands}.}
 \label{fig:wild_images}
 \end{figure}

\subsection{Extension to Images in the Wild}

We further present the hand reconstruction results on images in the wild. As shown in \cref{fig:wild_images}, our method performs well on real-life images taken from the RGB2Hands dataset \cite{wang2020rgb2hands}, which demonstrates the generalization ability of our approach. 
Recall that our method is designed to predict accurate mesh vertices positions and MANO parameters simultaneously in one forward pass. The MANO \textit{pose} parameter represents the joint rotations in axis-angle representation, which is useful for animating 3D hands in computer graphics. Accurate hand mesh vertices positions and MANO parameters for hands in the wild can facilitate different human-computer-interaction (HCI) applications.  



\subsection{Ablation Study}
To evaluate the effectiveness of the proposed modules in our framework, we conduct an ablation study on several variants of our method. The quantitative results of all variants are presented in Table \ref{table:ablation}.

\begin{figure}[!t]{}
 \captionsetup[subfigure]{labelformat=empty}
 \centering
{\includegraphics[width=0.35\textwidth]{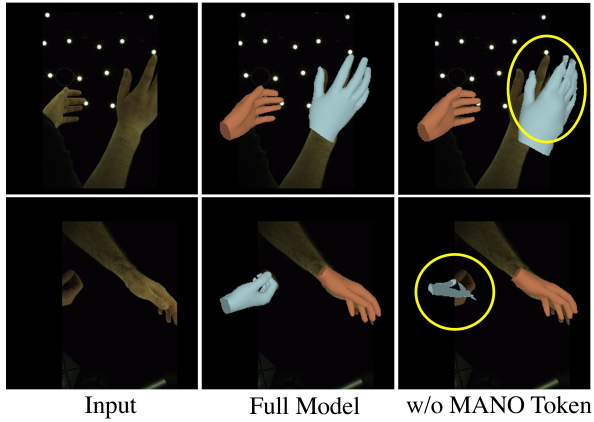}} 
 \caption{Ablation study on MANO tokens.}
 \label{fig:ablation_wo_mano}
 \end{figure}

\vspace{1mm}
\noindent \textbf{Effectiveness of MANO token.} Variant \textbf{A} (w/o MANO token) represents removing the MANO tokens. The transformer encoder only models the dependencies between the mesh vertex tokens and image features.  \cref{fig:ablation_wo_mano} show the qualitative comparison. Without hand prior information, the reconstructed mesh may be corrupted when parts of the hands are occluded. In contrast, our full model can generate reasonable meshes in these challenging situations.

\begin{table}[!t]
\centering
\resizebox{0.49\textwidth}{!}
{
\renewcommand{\arraystretch}{1.2}
\begin{tabular}{c|cccccc}
\specialrule{.1em}{0em}{0em}
       & MPJPE$\downarrow$ & MPVPE$\downarrow$ &  AUC$\uparrow$ & PROJ2D$\downarrow$ \\ \hline
\textbf{A}: w/o MANO token   & 8.87  & 9.09    & 0.818 & 6.85 \\ 
\textbf{B}: w/o Mesh-Align   & 8.79  & 9.01    &0.826 & 6.64\\ 
\textbf{C}:  Mesh-Align scale-16  & 8.73  & 8.97    &0.829 & 6.38\\ 
\textbf{D}:  Mesh-Align scale-64  & 8.77  & 8.99    &0.826 & 6.58\\ 
\textbf{E}: Regress MANO from $F_{gap}$ & 8.83 & 9.04 & 0.828 & 6.86\\
\textbf{F}: w/o asymmetric attention & 9.17 & 9.37 & 0.822 & 6.67\\
\textbf{G}: w/o auxiliary tasks & 9.11 & 9.28 & 0.823 & 6.78\\\hline
Full Model   & \textbf{8.65}   & \textbf{8.89}    &\textbf{0.832} & \textbf{6.22} \\ 
\specialrule{.1em}{0em}{0em}       
\end{tabular}
}
\caption{Comparisons of different variants of our method.}
\label{table:ablation}
\end{table}

\begin{figure}[!t]
 \centering
\includegraphics[width=0.48\textwidth]{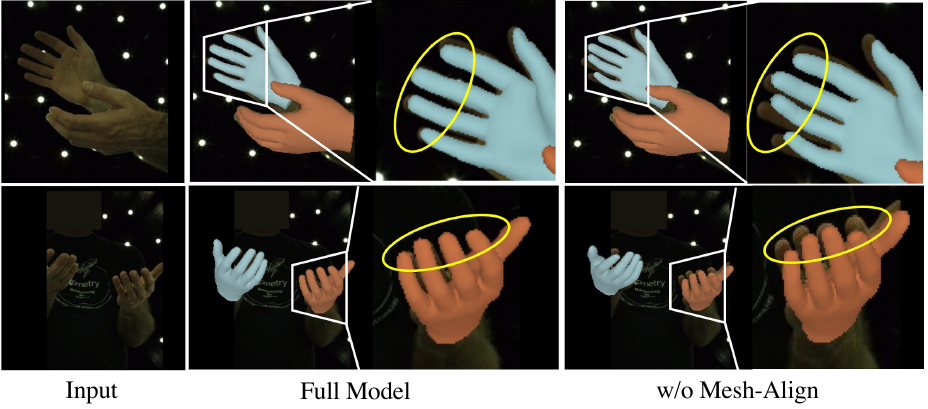}
 \caption{Ablation study on mesh alignment refinement.}
 \label{fig:ablation_wo_mesh_align}
 \end{figure}

\vspace{1mm}
\noindent \textbf{Effectiveness of Mesh Alignment Refinement.} Variant \textbf{B} (w/o Mesh-Align) represents removing the mesh alignment refinement module. The predictions generated by the Mesh-Mano interaction module are taken as the final outputs. Variant \textbf{C} and \textbf{D} denote extracting the single-scale image features at resolution 16 and 64, respectively. The performances degrade without using mesh align refinement or using single-scale image features. As shown in \cref{fig:ablation_wo_mesh_align}, our full model produces better-aligned results.

\vspace{1mm}
\noindent \textbf{Effectiveness of MMIB.} Parametric and non-parametric representations are widely used for many years. However, combining them is not trivial. In variant \textbf{E} (Regress MANO from $F_{gap}$), MMIB is responsible for updating vertex tokens. The MANO parameters are regressed from the global average pooling features $F_{gap}$ with simple fully connected layers rather than through MMIB. The performance gain of variant \textbf{E} is limited as shown in \cref{table:ablation}. On the other hand, we propose an asymmetric attention mask to exclude the mano-to-mesh attention. To verify the choice, variant \textbf{F} (w/o asymmetric attention mask) denotes removing the asymmetric attention mechanism. The performance degrades compared with our full model.

\vspace{1mm}
\noindent \textbf{Effectiveness of Pretraining on Auxiliary Tasks.} Variant \textbf{G} represents the backbone is pre-trained on ImageNet. Our full model performs better since pre-training on auxiliary tasks are essential to extract semantically-meaningful image features. The auxiliary 2D image conditions are generated from the ground-truth mesh of Interhand2.6M training split. We did not use extra datasets.



\begin{figure}[!t]{}
 \captionsetup[subfigure]{labelformat=empty}
 \centering
{\includegraphics[width=0.35\textwidth]{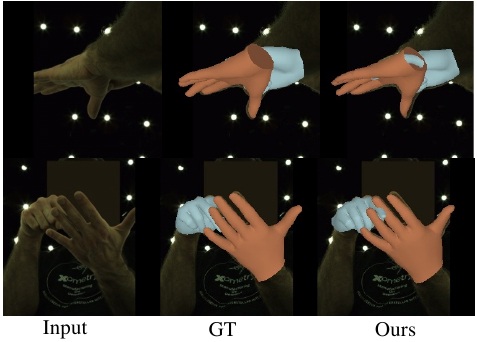}} 
 \caption{Failures cases caused by severe occlusions.}
 \label{fig:failure cases}
 \end{figure}

\section{Conclusion}
In this paper, we propose a novel approach MeMaHand for two-hand mesh reconstruction from a single image. The mesh-mano interaction module combines the merits of non-parametric and parametric representations. Then, the mesh alignment refinement module further rectifies the results with an explicit mesh alignment feedback loop. Extensive experiments on the InterHand2.6M benchmark demonstrate that our proposed MeMaHand is superior to both existing parametric and non-parametric methods. 

\vspace{1mm}
\noindent \textbf{Limitations.} Although our method generates promising results, it still fails in cases of severe occlusions. \cref{fig:failure cases} shows some failure cases, where inter-penetration occurs. Taking the physical plausibility into consideration will be our future work.



{\small
\bibliographystyle{ieee_fullname}
\bibliography{egbib}
}

\end{document}